\documentclass[conference]{IEEEtran}
\IEEEoverridecommandlockouts
\usepackage{cite}
\usepackage{amsmath,amssymb,amsfonts}
\usepackage{graphicx}
\usepackage{booktabs}
\usepackage{xcolor}
\usepackage{textcomp}
\usepackage{tikz}
\usetikzlibrary{arrows.meta, positioning, calc}

\begin{document}

\title{HAT Super-Resolution and a PARSeq+CLIP4STR Voting Ensemble \\ for Extreme In-the-Wild License Plate Recognition}

\author{
\IEEEauthorblockN{Karthik Sivarama Krishnan\IEEEauthorrefmark{1}\IEEEauthorrefmark{2}}
\IEEEauthorblockA{\IEEEauthorrefmark{1}Independent Researcher\\
\IEEEauthorrefmark{2}Ancestry\\
\texttt{ks7585@g.rit.edu}}
\and
\IEEEauthorblockN{Koushik Sivarama Krishnan\IEEEauthorrefmark{1}}
\IEEEauthorblockA{\IEEEauthorrefmark{1}Independent Researcher\\
\texttt{koushik.nov01@gmail.com}}
}

\maketitle

\begin{abstract}
We describe our entry to the ICIP 2026 Grand Challenge on Extreme In-the-Wild
License Plate Super-Resolution (XLPSR), which scored \textbf{9.73 wECR} on the
public validation leaderboard. The system pairs a Hybrid Attention Transformer
super-resolution (HAT) front-end with an ensemble of two scene-text recognisers
(PARSeq-S and CLIP4STR-B) and a confidence-weighted character-voting scheme
that abstains on uncertain positions. We treat XLPSR as a recognition task
gated by image legibility: the SR step exists to lift characters out of
sub-pixel territory, and the asymmetric scoring rule ($+2$/$-1$/$0$) is
exploited explicitly through abstention. Our pipeline runs in 1.7\,s per
sequence on RTX 3090 (max 2.7\,s, p99 2.4\,s), well under the
60\,s/sequence Docker budget.
\end{abstract}


\section{Introduction}
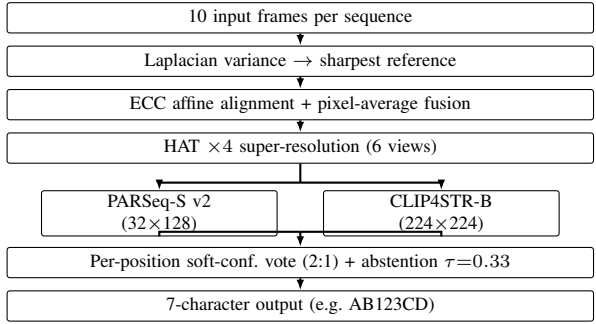
\begin{figure}[!t]
\centering
\begin{tikzpicture}[
  every node/.style={font=\scriptsize},
  box/.style={draw, rounded corners=1pt, align=center,
              minimum height=4mm, inner sep=2pt, text width=0.86\linewidth},
  smallbox/.style={draw, rounded corners=1pt, align=center,
                   minimum height=4mm, inner sep=1.5pt, text width=0.34\linewidth},
  arrow/.style={-{Latex[length=1.4mm]}, thick},
  line/.style={thick},
  node distance=1.6mm,
]
\node[box] (in) {10 input frames per sequence};
\node[box, below=of in] (var) {Laplacian variance $\to$ sharpest reference};
\node[box, below=of var] (ecc) {ECC affine alignment + pixel-average fusion};
\node[box, below=of ecc] (hat) {HAT $\times 4$ super-resolution (6 views)};
\coordinate (splitA) at ($(hat.south)+(0,-2.5mm)$);
\node[smallbox, anchor=north east] (par)  at ($(splitA)+(-3mm,-1mm)$) {PARSeq-S v2\\ (32$\times$128)};
\node[smallbox, anchor=north west] (clip) at ($(splitA)+(+3mm,-1mm)$) {CLIP4STR-B\\ (224$\times$224)};
\node[box, below=11mm of hat] (vote) {Per-position soft-conf.\ vote (2{:}1) + abstention $\tau{=}0.33$};
\coordinate (joinB) at ($(vote.north)+(0,+2mm)$);
\node[box, below=of vote] (out) {7-character output (e.g.\ AB123CD)};
\draw[arrow] (in) -- (var);
\draw[arrow] (var) -- (ecc);
\draw[arrow] (ecc) -- (hat);
\draw[line]  (hat.south) -- (splitA);
\draw[arrow] (splitA) -| (par.north);
\draw[arrow] (splitA) -| (clip.north);
\draw[line]  (par.south)  |- (joinB);
\draw[line]  (clip.south) |- (joinB);
\draw[arrow] (joinB) -- (vote.north);
\draw[arrow] (vote) -- (out);
\end{tikzpicture}
\caption{Pipeline. Ten sequence frames are ranked by Laplacian variance; five
aligned crops plus a temporally-fused crop are $4{\times}$ super-resolved by
HAT and passed through two scene-text recognisers whose per-position soft-max
outputs are combined by a confidence-weighted vote with $\tau{=}0.33$
abstention.}
\label{fig:pipeline}
\end{figure}

The XLPSR challenge provides 10 video frames per sequence of moving vehicles
with French license plates under extreme in-the-wild degradation: plates as
narrow as 12\,px, motion blur, JPEG compression, adverse lighting, and
occlusion. The task is to predict the plate text. The $+2$/$-1$/$0$
per-character scoring makes guessing in low-confidence regions
\emph{negative-EV} below $P(\textrm{correct}) = 1/3$, which we exploit
explicitly. Figure~\ref{fig:pipeline} summarises the pipeline; the
following sections describe each stage.

\section{Method}

\subsection{Multi-frame fusion}
Each sample in the blind-evaluation set is supplied as a temporal sequence of
already-cropped license-plate images; the public-validation set provides full
frames with per-frame bounding boxes which our pipeline crops first. After
that branching step the rest of the pipeline is identical: we score every
crop by Laplacian variance, pick the sharpest as a reference, align the
remaining crops to it with the Enhanced Correlation Coefficient (ECC)
algorithm under an affine motion model, and average them pixel-wise into a
fused crop. ECC + affine was chosen over optical-flow alternatives because
it is robust to the low-contrast conditions that dominate XLPSR crops,
requires no ground-truth flow supervision, and is fast enough for
CPU-side pre-processing. We carry forward six views per sequence into the
recogniser stack: the sharpest reference, up to four additional aligned
crops (in input frame order), and the fused crop.

\subsection{Super-resolution}
Each view is upscaled $4{\times}$ by Real-HAT-GAN-SRx4~\cite{chen2023hat}, a
Hybrid Attention Transformer with six RHAG blocks at embed dim 180, window 16,
overlap ratio 0.5, and a PixelShuffle upsampler. We use the public weights
without further fine-tuning. Lighter real-time variants such as our earlier
SwiftSRGAN~\cite{krishnan2021swiftsrgan} would suit tighter latency budgets,
but XLPSR's 60\,s/sequence budget lets us prioritise legibility over
throughput. SR is the single largest contributor to wECR in our ablation
(Table~\ref{tab:ablation}, $+2.00$ over the PARSeq+CLIP4STR no-SR baseline).

\subsection{Two-model OCR ensemble}
Both the super-resolved view and the original-resolution reference are
passed through two recognisers:

\begin{itemize}
\item \textbf{PARSeq-S v2}~\cite{bautista2022parseq}: 23.8\,M-parameter
ViT-Small with permuted autoregressive decoder, $32{\times}128$ input.
Fine-tuned from public weights for 10 epochs.
\item \textbf{CLIP4STR-B}~\cite{zhao2023clip4str}: 158\,M-parameter CLIP-ViT-B/16
backbone with cross-modal decoder, $224{\times}224$ input. Fine-tuned from
the public checkpoint with focal cross-entropy ($\gamma{=}2$),
label smoothing 0.1, differential LR (encoder $1{\times}10^{-5}$, decoder
$1{\times}10^{-4}$), EMA decay 0.999 and SWA over the last 30\% of training.
\end{itemize}

The two architectures give diversity at the backbone, input-resolution and
pretraining-corpus level (ImageNet-1K vs.\ CLIP's 400\,M image-text pairs)
while sharing the same permuted-AR decoder family. Every view therefore
yields two (predicted text, per-position soft-max confidence) tuples that
enter the voter.

\subsection{Character voting with abstention}
Predictions of the modal string length (preferring 7, the French SIV
format) are retained. For each character position we sum soft-max confidence
per class across all retained tuples, weighting PARSeq twice and CLIP4STR
once; the $2{:}1$ weighting is chosen from ablation (Table~\ref{tab:ablation},
$+0.04$~wECR over $1{:}1$). We then abstain (emit \texttt{\_}) when the
winning class's soft-max mass falls below $\tau = 0.33$, which matches the
$P(\textrm{correct})=1/3$ breakeven of the $+2/{-}1/0$ scoring rule under a
well-calibrated soft-max proxy. A small SIV-format correction maps
I$\to$J, O$\to$D, U$\to$V at the four letter positions of 7-character
plates, since the French SIV format forbids I, O and U to disambiguate them
from 1, 0 and V. Hard format constraints during voting hurt by
$-0.11$~wECR (Table~\ref{tab:ablation}) by filtering out correct edge-case
predictions and are therefore not used.

\section{Training Data}
We list every external resource exhaustively, as required by the challenge
rules. Sources are: (i) the XLPSR development set (39 sequences with ground
truth) used for OCR fine-tuning ($100{\times}$ oversampled) and validation
only; (ii) 200\,K self-generated synthetic French plates rendered with
OpenCV Hershey vector fonts (Apache-2.0 via OpenCV) and the community
FE-Schrift TrueType file (free for non-commercial use), degraded with a
BSRGAN~\cite{zhang2021bsrgan}-style shuffled pipeline (3--6 random ops per
sample, 40\% second-order cascade; difficulty split
$50/30/15/5\%$ very-hard/hard/medium/easy by plate width); and
(iii) public pretrained weights only:
PARSeq-S~\cite{bautista2022parseq}, CLIP4STR-B~\cite{zhao2023clip4str},
CLIP ViT-B/16~\cite{radford2021clip}, and HAT~\cite{chen2023hat}. No
private data and no manual annotation of public-validation or blind
sequences were used.

\section{Results}

Table~\ref{tab:ablation} traces our path from a no-SR baseline at 7.27 wECR
to the final 9.73 entry. Two headline findings drive the design:

\begin{itemize}
\item Super-resolution dominates: $+2.00$~wECR over the PARSeq+CLIP4STR
no-SR baseline, regardless of OCR choice.
\item Ensemble \emph{diversity} matters more than ensemble \emph{size}:
adding PARSeq+CLIP4STR over PARSeq alone cleanly adds $+0.31$~wECR, whereas
in preliminary trials a third recogniser (SVTRv2, CTC decoder) degraded the
ensemble --- an observation we return to below.
\end{itemize}

The dominance of super-resolution is a consequence of the extreme-scale
regime: at 12--20\,px plate widths, individual glyphs occupy only 2--3
pixels of stroke width, and OCR backbones trained on scene-text corpora have
effectively no signal until pixels are recovered. HAT lifts strokes above
this sub-pixel threshold, which is why the SR step contributes more than any
single recogniser choice or ensembling trick. The observed failure of the
SVTRv2 third member reflects a deeper principle: adding a model with a
different decoder family (CTC vs.\ permuted-AR) increases disagreement
without increasing correlated correctness, which is vote noise rather than
vote signal. Abstention at $\tau{=}0.33$ converts roughly 0.4 characters
per sequence from likely $-1$ errors into 0s, gaining $+0.11$~wECR by
calibrating exactly to the $+2/{-}1/0$ breakeven.

\begin{table}[!t]
\centering
\caption{Ablation on the public-validation set (higher wECR is better; max 10).}
\label{tab:ablation}
\begin{tabular}{lc}
\toprule
Configuration & wECR \\
\midrule
PARSeq + CLIP4STR (no SR) & 7.27 \\
PARSeq + HAT SR & 9.27 \\
PARSeq + CLIP4STR + HAT SR ($1{:}1$ vote) & 9.58 \\
PARSeq + CLIP4STR + HAT SR ($2{:}1$ vote) & 9.62 \\
\quad + hard format-constrained voting & 9.51 \\
\quad + abstention $\tau{=}0.33$ (final) & \textbf{9.73} \\
\bottomrule
\end{tabular}
\end{table}

\section{Inference Cost and Conclusion}
The full pipeline runs in $\sim$1.7\,s per sequence on RTX 3090 (max
2.7\,s, p99 2.4\,s; peak VRAM $\sim$3.2\,GB), leaving 22--35$\times$
headroom under the 60\,s/sequence limit. Careful super-resolution combined
with modest ensemble diversity and asymmetric-scoring-aware abstention
reaches 9.73 wECR on XLPSR without external data or proprietary models,
demonstrating that legibility recovery, not decoder capacity, is the
binding constraint on extreme-scale license-plate recognition.

\end{document}